\documentclass{esannV2}
\pdfoutput=1
\usepackage[dvips]{graphicx}
\usepackage[latin1]{inputenc}
\usepackage{amssymb,amsmath,array}
\usepackage{latexsym}
\usepackage{times}
\usepackage{booktabs}
\usepackage{paralist}

\usepackage{hyperref}
\hypersetup{
    colorlinks = true,
    linkcolor = {blue},
    urlcolor = {blue},
}
\begin{document}
\title{Neural Architecture Search for Sentence Classification with BERT}

\author{Philip Kenneweg$^1$, Sarah Schr{\"o}der$^1$ and Barbara Hammer$^1$
%
\thanks{We gratefully acknowledge funding by the BMWi (01MK20007E), from the MKW NRW in the project Bias aus KI-Modellen and by the BMBF (01IS19080 A).}
%
\vspace{.3cm}\\
%
1- Bielefeld University - Faculty of Technology \\
Inspiration 1, 33615 Bielefeld - Germany
}

\maketitle

\begin{abstract}
Pre training of language models on large text corpora is common practice in Natural Language Processing. Following, fine tuning of these models is performed to achieve the best results on a variety of tasks.
In this paper we question the common practice of only adding a single output layer as a classification head on top of the network. We perform an AutoML search to find architectures that outperform the current single layer at only a small compute cost. We validate our classification architecture on a variety of NLP benchmarks from the GLUE dataset. The source code is open-source and free (MIT licensed) software, available at \newline \url{https://github.com/TheMody/NASforSentenceEmbeddingHeads}.
\end{abstract}

\section{Introduction}
\label{sec:intro}
In the last couple of years, the transformer architecture pioneered by Vaswani et al. \cite{DBLP:journals/corr/VaswaniSPUJGKP17} has enabled large pre-trained neural networks to efficiently tackle previously difficult Natural Language Processing (NLP) tasks with relatively few training examples.
Pretrained transformer based language models, which produce a vectorized representation of a given text input of  arbitrary lengt, constitute a common tool which is deployed to facilitate fast transfer of knowledge.
These language models produce contextualized word embeddings that are easy to process for a multitude of different tasks by shallow neural networks or other machine learning tools - common technologies which are used on top of such embeddings are methods such as SVM, clustering algorithms and linear regression \cite{DBLP:journals/corr/abs-1902-00751,DBLP:journals/corr/abs-2005-14165,Radford2019LanguageMA}.
Most language models are pre-trained on a large corpus of text data (for example Wikipedia, Reddit, etc.) using a variety of different unsupervised pre-training objectives (Masked Language Modeling, Next Sentence Prediction, etc.) \cite{bert,bowman2015large}. 
Further, besides an often shallow layer on top of the embedding, many common architectures use a fine-tuning step on a specific task to achieve good performance \cite{DBLP:journals/corr/abs-1911-03437}.

In this paper we take a closer look at NLP architectures built on top of transformer models using subsequent fine tuning steps for its optimization. Based on current best practices, we aim for an automation of these optimization steps w.r.t\ architectural choices and training procedure. We achieve this by two contributions: first, we define a landscape of promising options for a classification head beyond the commonly used single layer neural network with a softmax activation. Second, we introduce an AutoML pipeline \cite{HE2021106622}, which enables us to automatically search these options to find the best possible classification architecture.


\label{sec:related_work}

At present, the most common approach to fine tuning a language model is to process the outputs of the transformer with a single layer neural network  \cite{DBLP:journals/corr/VaswaniSPUJGKP17,bert}. In the last years, more sophisticated fine tuning approaches have evolved to improve this baseline approach.
Some approaches focus on reducing the number of parameters that have to be fine tuned for each individual task. This can be done by adding specialized classification layers to the original network while keeping the original network weights frozen \cite{DBLP:journals/corr/abs-1902-02671}. 

Other approaches  take different avenues to address the challenges which occur while fine tuning . One challenge is that  the language model tends to forget the knowledge acquired by its extensive pre-training during fine tuning, or it over-fits on new data due to its inherent complexity and an often comparatively small size of the fine tuning data set.
The approach dubbed SMART \cite{DBLP:journals/corr/abs-1911-03437} addresses these shortcomings by introducing a smoothness inducing regularization technique and an optimization method, which prevents aggressive updating of the network weights.

In this paper we particularly introduce AutoML technologies, which enables us to investigate a variety of additional layers on top of the language model, while also fine tuning the weights of the underlying language model. To the best of our knowledge, this constitutes one of the first approaches in which the effect of additional classification architecture on the fine tuning process is extensively investigated.

\section{Neural Architecture Search Space}
\label{sec:req}

\begin{figure}[t]
    \centering
    \includegraphics[width=1\textwidth ]{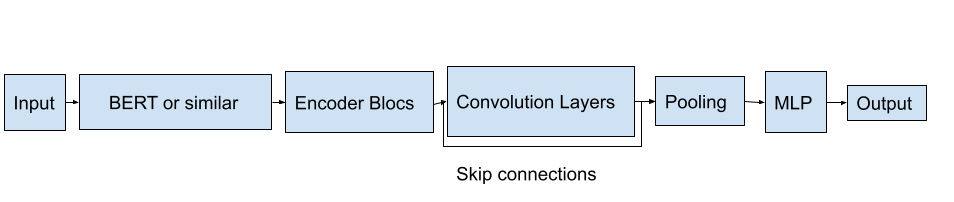}
    \caption{Rough outline of all possible options for the search space of the proposed AutoML pipeline. (Not all options are used all the time)}
    \label{fig:pipeline}
\end{figure}

The difficulty in finding a better classification head is caused by the fact that the transformer architecture is very powerful -- hence it is able to adapt to most classification tasks well but running the risk of forgetting. Hence a new and possibly complex classification head has to be able to add novel capacity to this capability, while not causing problems with vanishing gradients or other confounding factors for the base transformer.
With this in mind we determine a space of architectural choices, which constitute the possible search options in the neural architecture search which we are going to propose:
\begin{compactitem}
    \item pooling type (max, mean, [CLS])
    \item freeze base architecture (True,False)
    \item fully connected network (MLP); subchoices are:\\
    number of layers (1-5);
  number of neurons per layer (5-200))
    \item convolutional layer (True,False); subchoices are:\\
     number of heads (5-200);
       kernel size (3-11);
  number of layers (1-5);
      skip connections (True,False)
    \item encoder blocs (True,False); subchoices are:\\
     number of heads (1-16); number of layers (1-5)
\end{compactitem}
 A rough outline of the proposed Search Space for our neural architecture search can be seen in Figure \ref{fig:pipeline}.
The pooling type is specific to BERT and similar bidirectional language models upon which we focus in this work. During training, BERT is conditioned to produce a good embedding for classification in place of the [CLS] token output. Other options common in literature to feed the output into the fully connected network are mean and max pooling over all output tokens.
In between the dense layers of the MLP and the convolutional layers the activation function used is ReLU. 5 layers and 200 neurons for any type of layer was deemed the maximum to keep the parameter count of the new classification head to a reasonable amount (below 1\% of the original BERT architecture).
The convolutional layers were padded with zeroes to keep the input and output dimensions the same.

The search space spans $7.5e7$ different possible combination options if one assumes 10 different search options for the number of neurons per layer and number of heads during convolution.
The search strategy used for our AutoML is Bayesian Optimization \cite{bayesopt} with Hyperband Scheduling \cite{li2017hyperband}, as a particularly promising technology in the domain of architecture optimization.
We refer to the output of this procedure, i.e., the best configuration found by this approach in a specific traning task, as $BERT tuned$. 

\section{Experimental Approach}
\label{sec:experiments}

In this section we detail our experimental design to investigate the effects of different architectures for Natural language classification tasks.
We utilize AutoGluon \cite{abohb} and the BertHugginface library \cite{wolf2020huggingfaces} for implementation and the pre-trained Bert model ('bert-base-uncased') for all experiments.  

\subsection{Datasets}

The Glue dataset by Wang et al. \cite{wang2019glue} is a collection of various popular datasets in NLP, and it is widely used to evaluate common natural language processing capabilites. All datasets used are the version provided by tensorflow-datasets 4.0.1. We also evaluate all approaches on the $small$ datasets. These are the same datasets as described below, only with the training data set size reduced to 500 randomly drawn samples. This scaling enables us to judge the capability of the architectural choices to adapt to new tasks based on a small number of training data only.
More specifically, we use the Corpus of Linguistic Acceptability \emph{cola}, the Stanford Sentiment Treebank \emph{sst2}, the 
 Microsoft Research Paraphrase Corpus \emph{mrpc}, Recognizing Textual Entailment \emph{rte},
 and the Multi-Genre Natural Language Inference Corpus 
  \emph{mnli}.
  




\begin{table}[t]
  \centering
  \caption{The final classification architectures found for the GLUE datasets.}
  \label{Fig:gluearch}
  \begin{tabular}{c cccccc}
    \toprule
method & sst2 & cola & mrpc & mnli & rte & qqp  \\
 \cmidrule(r){1-1} \cmidrule(l){2-7} 
pooling & mean & [CLS] & [CLS] & max & [CLS] & max  \\
number linear layers & 5 & 1 & 4 & 3 & 1 & 1 \\
hidden dim linear layers & 50 & - & 74 & 117 & - & -   \\
number conv layers & 0 & 0 & 4 & 3 & 0 & 2  \\
number heads conv & - & - & 107 & 9 & - & 159  \\
kernel size & - & - & 7 & 11 & - & 7  \\
skip connection & - & - & True & True & - & True  \\
number attention layers & 1 & 0 & 4 & 0 & 0 & 0  \\
number attention heads & 4 & - & 16 & - & - & - \\
    \bottomrule
  \end{tabular}
\end{table}

\begin{table}[b]
  \centering
  \caption{Classification accuracies, on the sst2, cola, mrpc, rte, qqp and mnli datasets. 
  Improvements are marked in \textbf{bold}.}
  \label{Fig:glue}
  \begin{tabular}{c cccccc c}
    \toprule
method & sst2 & cola & mrpc & mnli & rte & qqp & average \\
 \cmidrule(r){1-1} \cmidrule(l){2-7} \cmidrule(l){8-8}
$BERT base$ &  0.925 & 0.831  & 0.821 & 0.829 & 0.700 & 0.899 & 0.833 \\
$BERT tuned$ & \textbf{0.930} &  0.831 & \textbf{0.860} & \textbf{0.835} & 0.700 & \textbf{0.900} &  \textbf{0.842} \\

    \bottomrule
  \end{tabular}
\end{table}

\subsection{Baseline and Implementation Details}

As a Baseline comparison we evaluate BERT with a single dense layer used for classification on top and a learning rate of 2e-5, henceforth referenced to as $BERTbase$. The pooling operation used is [CLS]. Mean and max pooling were also tried but produced inferior results. These values are taken from the original paper \cite{bert} and present good values for a variety of classification tasks.

\label{sec:impl}
All models are trained using a cosine decay of their learning rate with warm starting.
All models are trained for 5 epochs on the glue datasets and for 10 epochs on the $small$ glue datasets.
Batch size used for training was 32 and the Adam optimizer with betas (0.9,0.999) and epsilon 1e-08 was used.

\section{Results}

For our experiments, the 6 glue tasks cola, sst2, mnli, mrpc, rte and qqp are considered, following common evaluation approaches from literature. 
As can be seen in Table \ref{Fig:glue} $BERTtuned$ outperforms $BERTbase$ by a solid margin. On the cola as well as on the rte task the architecture search seems to have yielded little improvement, but on the mrpc task the accuracy was improved more significantly by 4\%.
On average the accuracy was improved by 0.9\%. Larger accuracy improvement seems to correlate to more extensive changes to the classification head architecture as displayed in Table \ref{Fig:gluearch}. The architectures found were using multiple convolutional layers and multiple encoder blocks, but never more than 2 dense layers in the MLP. All optimal architectures are adding skip connections if a convolutional layer is added. For all results (thus not in displayed in Table \ref{Fig:gluearch}) the base layer was not frozen (freeze Base = False); this indicates,  that the underlying transformer architecture is also adjusted towards the new tasks and not simply replaced by a new classification head.

In Tables \ref{Fig:gluesmall} and \ref{Fig:gluesmallarch} we see according results of the architecture search for the $small$ datasets. Here we see even more significant improvements than on the full datasets. This can be explained by the overall lower performance of the models and hence larger  margin for improvement. The classification architectures found outperform the baseline in all cases. In this case the architectures found generally had less parameters to train than in the case of the full datasets. In all cases either only a convolutional layer, or an encoder block was added on top of the network, never both. On average the classification performance was improved by 3\% over $BERTbase$. In comparison to the full datasets we see that even with fewer training examples it is still possible to upgrade the performance of $BERTbase$; thereby, the additional architecture has fewer additional parameters to avoid overfitting on the training data.



\begin{table}[t]
  \centering
  \caption{The final classification architectures found for the GLUE $small$ datasets.}
  \label{Fig:gluesmallarch}
  \begin{tabular}{c cccccc}
    \toprule
method & sst2 & cola & mrpc & mnli & rte & qqp  \\
 \cmidrule(r){1-1} \cmidrule(l){2-7} 
pooling & max & [CLS] & mean & max & mean & [CLS] \\
number linear layers & 1 & 1 & 2 & 2 & 1 & 2 \\
hidden dim linear layers & - & - & 60 & 172 & - & 122   \\
number conv layers & 0 & 0 & 2 & 2 & 5 & 1  \\
number heads conv & - & - & 90 & 75 & 14 & 43  \\
kernel size & - & - & 7 & 11 & 5 & 11  \\
skip connection & - & - & True & False & True & False  \\
number attention layers & 1 & 0 & 0 & 0 & 0 & 0  \\
number attention heads & 8 & - & - & - & - & - \\
    \bottomrule
  \end{tabular}
\end{table}

\begin{table}[b]
  \centering
  \caption{Classification accuracies, for the $small$ datasets. 
  Improvements are marked in \textbf{bold}.}
  \label{Fig:gluesmall}
  \begin{tabular}{c cccccc c}
    \toprule
method & sst2 & cola  & mrpc  & mnli  & rte & qqp & average \\
 \cmidrule(r){1-1} \cmidrule(l){2-7} \cmidrule(l){8-8}
$BERT base$ &  0.835 & 0.761  & 0.733 & 0.463 & 0.566 & 0.736 & 0.682 \\
$BERT tuned$ & \textbf{0.869} &  \textbf{0.763} & \textbf{0.752} & \textbf{0.529} & \textbf{0.606} & \textbf{0.751} & \textbf{0.712} \\
    \bottomrule
  \end{tabular}
\end{table}



\section{Conclusions}
\label{sec:conclusion}

In this paper we presented a classification architecture for the BERT sentence embedder which improves the classification performance on a variety of datasets. 
In comparison to other approaches we do not modify the underlying transformer architecture or provide additional regularization, but rather append more complex network heads to the BERT network. This is agnostic towards the underlying architecture as well as task and so is more flexible in improving classification performance among a variety of architectures and tasks than many other approaches.

The exact contribution of the classification architectures found in comparison to a single layer is of interest and merits further research.
Possible further research also includes other options for the search space of the NAS, as well as different transformer based language models. 

\noindent
The source code is available at \newline \url{https://github.com/TheMody/NASforSentenceEmbeddingHeads}, \newline It is open-source and free (MIT licensed) software.

\section*{Acknowledgements}
We gratefully acknowledge funding by the BMWi (01MK20007E), from the MKW NRW in the project Bias aus KI-Modellen and by the BMBF (01IS19080 A).


\begin{footnotesize}


\bibliographystyle{unsrt}
\bibliography{references}

\end{footnotesize}


\end{document}